\documentclass[11pt,a4paper]{article}
\usepackage[hyperref]{emnlp2020}
\usepackage{algorithm}%
\usepackage{algcompatible}
\usepackage{times}
\usepackage{latexsym}
\usepackage{amsmath}
\usepackage{multirow}
\usepackage{xcolor}
\usepackage{mathtools}
\usepackage{microtype}
\usepackage{paralist}
\usepackage{xspace}
\usepackage{booktabs}
\usepackage{url}

\DeclarePairedDelimiter\evaluat{.}{\rvert}

\aclfinalcopy %
\def\aclpaperid{884} %

\newcommand{\draftonly}[1]{#1}
\renewcommand{\draftonly}[1]{} %

\newcommand{\reasoning}{commonsense reasoning\xspace}

\newcommand{\gdaug}{$\textsc{G-DAug}^{\textbf{c}}$}

\newcommand{\winogrande}{\textsc{WinoGrande}\xspace}
\newcommand{\codah}{\textsc{Codah}\xspace}
\newcommand{\comqa}{\textsc{CommonsenseQA}\xspace}
\newcommand{\gpt}{\textsc{GPT-2}\xspace}
\newcommand{\bert}{\textsc{BERT}\xspace}
\newcommand{\roberta}{\textsc{RoBERTa}\xspace}
\newcommand{\backt}{\textsc{BackTranslation}\xspace}

\title{
Generative Data Augmentation for Commonsense Reasoning}

\author{
    Yiben Yang$^\spadesuit$ $\quad$ 
	Chaitanya Malaviya$^{\dagger\Pi}$ $\quad$
    Jared Fernandez$^{\spadesuit\heartsuit}$ $\quad$  
	Swabha Swayamdipta$^{\dagger}$ $\quad$ \\
	\bf Ronan Le Bras$^\dagger$ $\quad$
	Ji-Ping Wang$^\spadesuit$ $\quad$
	Chandra Bhagavatula$^{\dagger}$
	Yejin Choi$^{\dagger\diamondsuit}$	
	Doug Downey$^{\spadesuit\dagger}$ $\quad$  \\\\
    $^\spadesuit$Northwestern University, Evanston, IL, USA \\
	$^\dagger$Allen Institute for Artificial Intelligence, Seattle, WA, USA \\
	$^\Pi$ University of Pennsylvania, Philadelphia, PA, USA \\
	$^\heartsuit$ Carnegie Mellon University, Pittsburgh, PA, USA \\
	$^\diamondsuit$University of Washington, Seattle, WA, USA \\
	{\tt \{yiben.yang@,jared.fern@u.,jzwang@\}northwestern.edu} \\
	{\tt \{chaitanyam,swabhas,ronanlb,chandrab,yejinc,dougd\}@allenai.org}  
}

\date{}

\begin{document}
\maketitle
\begin{abstract}

Recent advances in \reasoning depend on large-scale human-annotated training sets to achieve peak performance. 
However, manual curation of training sets is expensive and has been shown to introduce annotation artifacts that neural models can readily exploit and overfit to. 
We propose a novel generative data augmentation technique, \textbf{\gdaug}, that aims to achieve more accurate and robust learning in a low-resource setting. 
Our approach generates synthetic examples using pretrained language models, and selects the most informative and diverse set of examples for data augmentation. 
On experiments with multiple commonsense reasoning benchmarks, \gdaug\ consistently outperforms existing data augmentation methods based on back-translation, establishing a new state-of-the-art on \winogrande, \codah, and \comqa, and also enhances out-of-distribution generalization, proving to be more robust against adversaries or perturbations.
Our analysis demonstrates that \gdaug\ produces a diverse set of fluent training examples, and that its selection and training approaches are important for performance. 
\end{abstract}

\section{Introduction}
\label{sec:intro}

\begin{figure}[t!]
\centering
  \includegraphics[width=1\columnwidth]{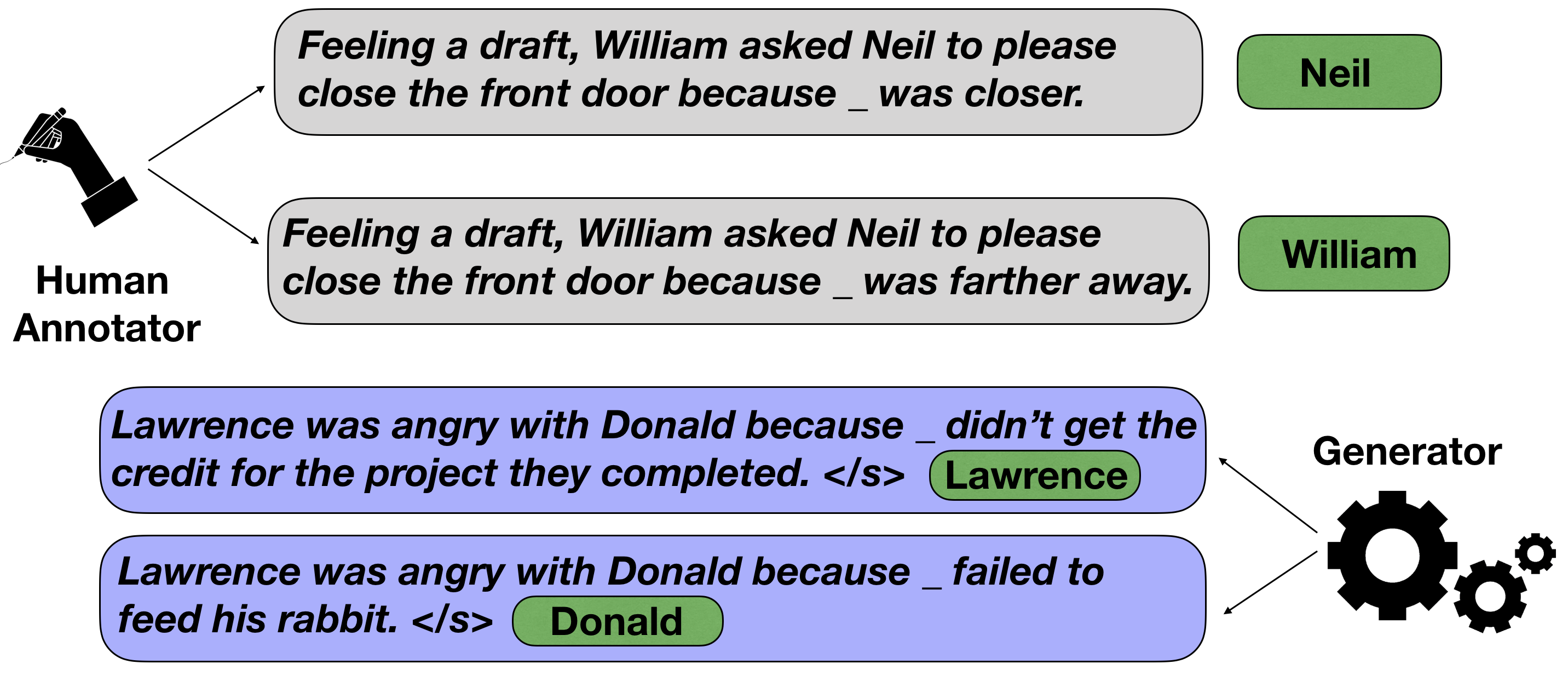}
  \caption{Example of a selected high-quality generated example compared to a human-authored example from the \winogrande dataset. Composing commonsense questions can require creativity.}
  \label{fig:winogrande_gen}
\end{figure}

While recent advances in large-scale neural language models \cite{devlin-etal-2019-bert, liu2019roberta, radford2019language, 2019t5} have led to strong performance on several commonsense reasoning benchmarks \cite{talmor-etal-2019-commonsenseqa, hekcqa_aaai20, sakaguchi2019winogrande}, their accuracy by and large depends on the availability of large-scale human-authored training data. 
However, crowdsourcing examples at scale for each new task and domain can be prohibitively expensive.  
Moreover, 
human-authored data has been shown to exhibit annotation artifacts %
\cite{gururangan-etal-2018-annotation,Agrawal2018DontJA,schwartz-etal-2017-effect}, 
leading to models with considerably weaker performance on out-of-distribution samples \cite{jia-liang-2017-adversarial,belinkov2017synthetic,iyyer-etal-2018-adversarial}.

A candidate solution that has shown promise in other tasks, such as reading comprehension, is to augment a human-authored training set with a large set of synthetically-generated examples  \cite{zhou2017neural,du2017learning,zhao-etal-2018-paragraph}.  But, generating synthetic examples for commonsense reasoning poses a unique challenge.
In reading comprehension, for instance, the goal of data augmentation is to generate questions that are directly answerable by a given reference passage.
In contrast, answering commonsense questions relies on commonsense notions that are seldom stated explicitly \cite{gordon2013reporting,forbes-choi-2017-verb}, and authoring such questions can require creativity (see Figure \ref{fig:winogrande_gen}). 
Based on promising evidence from previous work \cite{Yang2018ExtractingCP,Trinh2018ASM,Bosselut2019COMETCT, davison-etal-2019-commonsense}, we hypothesize that pretrained language models, such as \gpt \cite{radford2019language}, capture some common sense expressed implicitly in their pretraining corpus.
Could questions generated by such models serve as helpful training data?
In this work, we explore this question through \textbf{G}enerative \textbf{D}ata \textbf{Aug}mentation for {\bf c}ommonsense reasoning (\textbf{\gdaug};  \S\ref{sec:gdaug}): a novel framework for augmenting training data with \textit{diverse} and \textit{informative} synthetic training examples to improve both in-distribution performance and out-of-distribution generalization of commonsense reasoning models.\footnote{https://github.com/yangyiben/G-DAUG-c-Generative-Data-Augmentation-for-Commonsense-Reasoning}

Although a generative model allows us to produce large pools of synthetic training examples, the generated examples may be noisy or redundant.
To ensure that we use the most informative examples for augmentation, we introduce data selection methods based on influence functions \cite{Koh2017UnderstandingBP} and a heuristic to maximize the diversity of the generated data pool. 
Finally, we propose an effective two-stage training scheme for augmentation with synthetic data.
In experiments across multiple commonsense benchmarks, we show that \gdaug\ can mitigate the expense and brittleness resulting from large training sets for \reasoning tasks. 

To summarize, our contributions include: 
\begin{compactenum}
    \item \gdaug, a generative data augmentation framework for commonsense reasoning (\S\ref{sec:gdaug}), %
     \item novel selection methods that identify \textit{informative} and \textit{diverse} synthetic training examples from the generated pool (\S\ref{sec:selection}),
    \item experiments showing that \gdaug\ improves in-distribution performance, achieving a $1$--$4$\% average absolute gain across four commonsense reasoning data sets and state-of-the-art results on the \winogrande \cite{sakaguchi2019winogrande}, \comqa \cite{talmor-etal-2019-commonsenseqa}, and \codah \cite{Chen2019CODAHAA} benchmarks, and also improves model robustness in terms of resistance to adversarial attacks \cite{jin2019bert} and accuracy on perturbed evaluation sets (\S\ref{ref:sec-exp}), and
    \item a comprehensive analysis of the factors that influence \gdaug's performance (\S\ref{sec:analysis}).
\end{compactenum}

\begin{figure}[t!]
\centering
  \includegraphics[width=1\columnwidth]{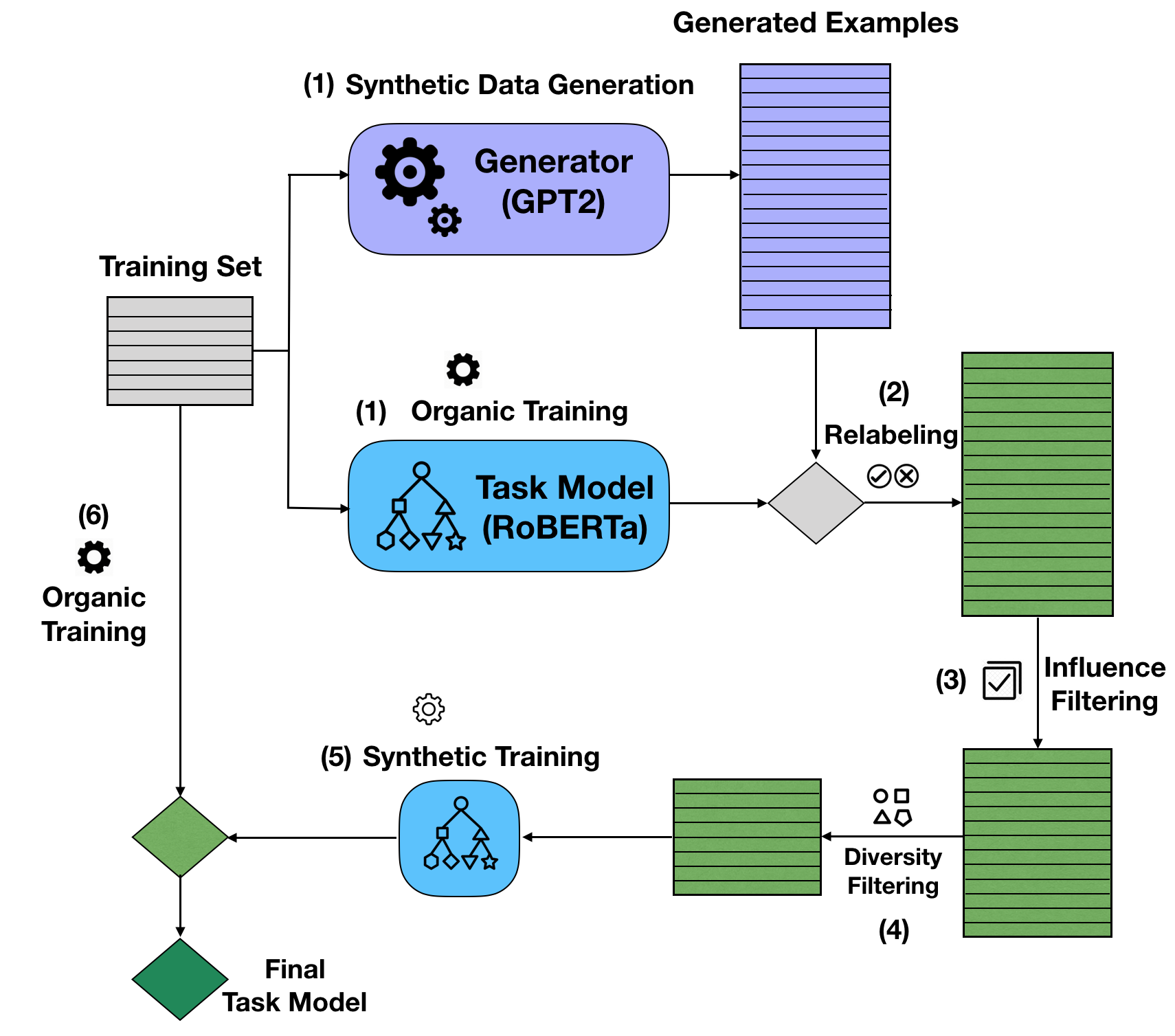}
  \caption{Illustration of the \gdaug\ process: (1) generate synthetic data and train a task model, (2) relabel the generated data using the task model, (3) filter the generated data based on estimated influence scores, (4) further select a subset based on a diversity-maximizing heuristic, (5) train a new task model using the filtered generations (\textit{synthetic training}), and (6) further train this model using the original training data (\textit{organic training}).}
  \label{fig:gdaug}
\end{figure}
\section{\gdaug}
\label{sec:gdaug}

We now describe our framework for \textbf{G}enerative \textbf{D}ata \textbf{Aug}mentation for \textbf{C}ommonsense Reasoning (\textbf{\gdaug}).
Figure \ref{fig:gdaug} shows an overview of the approach.  We describe \gdaug 's data generation procedure (steps 1 and 2 in the figure) in this section, and cover the data selection and training components (steps 3-5) in \S\ref{sec:selection}.

\subsection{Synthetic Training Data Generation}
\label{sec:data-gen}

We will use multiple choice question answering as a running example to describe synthetic data generation.
Formally, consider a dataset of $N$ questions $\mathcal{D}=\{(\mathbf{Q}^{i},\mathcal{C}^{i},y^{i}): i = 1,2,...,N\}$,  where $\mathbf{Q}^{i}$ is a sequence of words denoting the $i^{th}$ question, $\mathcal{C}^{i}=\{ \mathbf{C}^{i}_{j}:j = 1,2,...,K \}$ is the corresponding choice set with $K$ choices which are word sequences as well,
and a ground truth label $y^i \in \{1,2,...,K\}$. 
We denote the \textbf{answer} as $\mathbf{C}_{y^i}^i$ %
and the \textbf{distractors} as $\mathbf{C}_{j \neq y^i}^i$s.

Our text generators are pretrained generative language models, finetuned to maximize the log-likelihood of a sequence of text $\mathbf{W}$, $ \mathcal{L}_W(\boldsymbol{\theta}) =  \sum_{t=1}^{T} \log P(w_t|\mathbf{W}_{1:t-1};\boldsymbol{\theta})$,
where $\mathbf{W}_{1:t-1}$ denotes a subsequence of $\mathbf{W}$
and  $\boldsymbol{\theta}$ denotes the model parameters.\footnote{ $W_{1:0}$ denotes an empty sequence}
Below, we describe how we use variations of this objective to finetune different LMs to generate questions, answers and distractors.\footnote{
Specific modifications for other tasks, e.g. textual entailment, are discussed in  Appendix \ref{app:datasets}.}

\paragraph{Generating Synthetic Questions} 
To train our question generator, we finetune the LM on the training question set $\{\mathbf{Q}^{i}\}$ to optimize the language modeling objective: $\mathcal{L}_q(\boldsymbol{\theta}_{q}) = \sum_{i=1}^N\log P(\mathbf{Q}^{i};\boldsymbol{\theta}_{q}),$
where $\boldsymbol{\theta}_{q}$ denotes the parameters of the question generator. After finetuning, we generate new questions with nucleus sampling \cite{holtzman2019curious}, which is suitable for generating long-form text.

\paragraph{Generating Synthetic Answers and Distractors} 
To generate choice sets, we independently finetune two separate generative LMs, one for answers and the other for distractors.        %
The answer and distractor generators are trained to maximize the conditional log-likelihood of the answer and the distractors, respectively, given the question.
Mathematically, we optimize both $\mathcal{L}_a(\boldsymbol{\theta}_a) = \sum_{i=1}^N \log P(\mathbf{C}^{i}_{y^{i}}|\mathbf{Q}^{i};\boldsymbol{\theta}_a)$ and $\mathcal{L}_d(\boldsymbol{\theta}_d) = \sum_{i=1}^N\sum_{j\neq y^{i} } \log P(\mathbf{C}^{i}_{j}|\mathbf{Q}^{i};\boldsymbol{\theta}_d)$,
where $\boldsymbol{\theta}_{a}$ and  $\boldsymbol{\theta}_{d}$ denote the parameters of the answer and distractor generators, respectively.
For answers, we use nucleus sampling with low temperature (for long answers) or greedy decoding (for short answers). To encourage diversity across generated distractors, we use nucleus sampling without temperature for these.

\paragraph{Data Relabeling.} 

Our choice of generative LMs naturally defines labels for the synthetic choice sets. 
Alternatively, we consider using a supervised task model trained on the original training set, to {\em relabel} a candidate pool of synthetic answers and distractors. 
This is similar to treating the synthetic questions as unlabeled data and applying self-training. 
The utility of this self-training can be task-dependent; in our experiments, we used validation performance to determine whether or not to relabel our synthetic training data.

\section{Synthetic Data Selection and Training}
\label{sec:selection}

The above generation method can produce a large pool of examples, but training on all of them would be computationally expensive and might harm performance due to noisy generations. Here, we propose three data selection methods aimed at choosing more effective training examples from the generated pool (\S\ref{sec:data-aug:select}).
Further, we outline a simple staged training procedure (\S\ref{sec:data-aug:training}) to mitigate the negative impact from noise in the synthetic data.

\subsection{Selecting High-quality and Diverse Synthetic Examples }
\label{sec:data-aug:select}
A randomly sampled synthetic dataset may contain examples that are similar to one another, along with low-quality generations \cite{holtzman2019curious}. We refer to such a random selection approach as \textbf{\gdaug-Rand}. We hypothesize that a diverse and high-quality synthetic set would benefit the task model more. 
We present three data selection algorithms that target quality, diversity and a combination of both.  

\paragraph{Filtering with Influence Functions.}
We hypothesize that filtering out detrimental synthetic training examples can boost downstream performance \cite{bras2020adversarial}.
A given training example $x$ is considered {\em detrimental} if including $x$ in the training set results in a higher generalization error, approximated by validation loss, i.e.:
\begin{align*}
&\mathcal{L}(\mathcal{X},\boldsymbol{\theta})=\frac{1}{|\mathcal{X}|}\sum_{x_i\in\mathcal{X}}l(x_i,\boldsymbol{\theta}),\\
&\mathcal{L}(\mathcal{X}_{val},\hat{\boldsymbol{\theta}}(\mathcal{X}_{tr}\cup \{x\}))-\mathcal{L}(\mathcal{X}_{val},\hat{\boldsymbol{\theta}}(\mathcal{X}_{tr}))>0.
\end{align*}
This would naively require retraining the model with $x$, which is computationally prohibitive.
Fortunately, the validation loss change can be efficiently approximated through the use of influence functions \cite{Atkinson1983ResidualsAI,Koh2017UnderstandingBP}. While previous work focuses on removing or perturbing 
existing training examples \cite{Koh2017UnderstandingBP,Wang2018DataDO}, we use influence functions to estimate the effect of including a novel synthetic example.

The main result from previous work \citep{Atkinson1983ResidualsAI,Koh2017UnderstandingBP} tells us that the influence of upweighting a training example $x$ by some small $\epsilon$ on the model parameters $\hat{\boldsymbol{\theta}}$ with the corresponding parameter space $\Theta$ is given by:
\begin{align*}
&\hat{\boldsymbol{\theta}}_{\epsilon,x}=\underset{\boldsymbol{\theta}\in \Theta}{\mathrm{argmin}}\ \epsilon l(x,\boldsymbol{\theta}) + \frac{1}{\sum_{i=1}^{N} w_i}  \sum_{i=1}^{N} w_i l(x_i,\boldsymbol{\theta}) \\
&\mathcal{I}_{up,params}(x):= \evaluat*{\frac{d\hat{\boldsymbol{\theta}}_{\epsilon,x}}{d\epsilon}}_{\epsilon=0}=-H^{-1}_{\hat{\boldsymbol{\theta}}}\nabla_{\boldsymbol{\theta}}l(x,\hat{\boldsymbol{\theta}}),
\end{align*}
where $w_i$ is weight for the training example $x_i$ and $H_{\hat{\boldsymbol{\theta}}}$ is the Hessian evaluated at $\hat{\boldsymbol{\theta}}$. The above result is a slight generalization of \citet{Koh2017UnderstandingBP}, but it is straightforward to generalize their proof to the weighted empirical risk case. Then, we apply the chain rule to get the influence of upweighting $x$ on the validation loss:
\begin{align*}
\label{up_inf_loss}
&\mathcal{I}_{up,loss}(x):= \evaluat*{\frac{d\mathcal{L}(\mathcal{X}_{val},\hat{\boldsymbol{\theta}}_{\epsilon,x})}{d\epsilon}}_{\epsilon=0}\\
&= \nabla_{\boldsymbol{\theta}}\mathcal{L}(\mathcal{X}_{val},\hat{\boldsymbol{\theta}})^\top \mathcal{I}_{up,params}(x).
\end{align*}
Note that $\mathcal{L}(\mathcal{X}_{tr},\boldsymbol{\theta})$ can be rewritten as the following weighted average form to incorporate a new training example $x_{new}$:
\begin{align*}
    \mathcal{L}(\mathcal{X}_{tr},\boldsymbol{\theta}) =& \frac{1}{\sum_{i=1}^{N+1}w_i}\sum_{i=1}^{N+1} w_i l(x_i,\boldsymbol{\theta}),
 \end{align*}
 where $ w_i = 1 \forall{i \neq N+1}$, $w_{N+1} = 0$ and $x_{N+1} = x_{new}$. 
Adding the new training example $x_{new}$ is equivalent to upweighting $x_{N+1}$ by $\frac{1}{N}$:
\begin{align*}
&\mathcal{L}(\mathcal{X}_{tr}\cup \{x_{new}\},\boldsymbol{\theta})\propto\frac{1}{N}l(x_{N+1},\boldsymbol{\theta}) \\
&+ \frac{1}{\sum_{i=1}^{N+1}w_i}\sum_{i=1}^{N+1}w_{i}l(x_i,\boldsymbol{\theta}). 
\end{align*}

Applying the influence function $\mathcal{I}_{up,loss}(x)$, we obtain the following linear approximation of the validation loss change upon adding the training example $x_{new}$:
\begin{align*}
\nonumber
&\mathcal{L}(\mathcal{X}_{val},\hat{\boldsymbol{\theta}}(\mathcal{X}_{tr}\cup \{x_{new}\}))-\mathcal{L}(\mathcal{X}_{val},\hat{\boldsymbol{\theta}}(\mathcal{X}_{tr}))\\
&\approx \frac{1}{N}\mathcal{I}_{up,loss}(x_{new}). 
\end{align*}
We adopt the stochastic estimation method described in \citet{Koh2017UnderstandingBP} to efficiently compute $\mathcal{I}_{up,loss}$. Detrimental synthetic data will have $\frac{1}{N}\mathcal{I}_{up,loss}>0$.

Another distinction between our approach and \citet{Koh2017UnderstandingBP} is that they compute the influence of a single training example on a single \textit{test} example, whereas
we estimate influence of a synthetic training example on all validation examples at once, which makes our approach scalable to large pools of synthetic data.
Our approach, referred to as \textbf{\gdaug-Influence}, filters out detrimental synthetic data (i.e., the examples that have a positive estimated influence on the validation loss).

\paragraph{Selecting Diverse Examples.} 
While \gdaug-Influence promotes training data quality, it ignores diversity; we hypothesize that better diversity can provide a more reliable training signal. 
We propose a simple greedy algorithm that iteratively selects a synthetic training example from the pool that maximizes a diversity measure.  Here, we use a simple measure of diversity equal to the number of unique unigrams in the selected training set.  Surprisingly, preliminary experiments with a more sophisticated diversity method based on embedding distance did not improve results (see Appendix \ref{app:emb} for details).
We refer to this approach as \textbf{\gdaug-Diversity} (see Algorithm \ref{alg:g-daug-diversity}).
 \begin{algorithm}[H]
   \caption{\gdaug-Diversity}
   \label{alg:g-daug-diversity}
 \begin{algorithmic}
   \STATE {\bfseries Input:} Synthetic data pool $\mathcal{D}_{pool}$, Target size $N$
   \STATE {\bfseries Output:} Synthetic dataset 
       \STATE Initialization: $\mathcal{D}_{synthetic} \xleftarrow{} \{\}$ 
     \REPEAT
     \STATE $x_{max} = \mathrm{argmax}_{x \in D_{pool}} \text{\#n-grams}(\mathcal{D}_{synthetic}$ \\
     \hspace{0.7in}$\cup \{x\}) -  \text{\#n-grams}(\mathcal{D}_{synthetic} )$
     \STATE Add $x_{max}$ to $\mathcal{D}_{synthetic}$
     \STATE Remove $x_{max}$ from $\mathcal{D}_{pool}$
  \UNTIL{$|\mathcal{D}_{synthetic}| = N$ }
  \STATE {\bf return} $\mathcal{D}_{synthetic}$
 \end{algorithmic}
 \end{algorithm}

\paragraph{Combining Influence Filtering and Diversity Maximization}
\gdaug-Influence and \gdaug-Diversity have complementary benefits---the former aims at improving the quality of individual examples by filtering out detrimental ones, and the latter is designed to compose a diverse training set but does not consider quality.  To reap both benefits, we propose a combined selection technique, \textbf{\gdaug-Combo}, that first filters the data using \gdaug-Influence, then selects examples according to \gdaug-Diversity.

\subsection{ Training with Synthetic Data}
\label{sec:data-aug:training}

In traditional data augmentation, new data is usually mixed with the original training examples to create an augmented training set \cite{wei-zou-2019-eda,kafle-etal-2017-data}. 
However, when augmenting with data produced using a generative model, label noise can be detrimental to learning \cite{kafle-etal-2017-data}. 
Moreover, the generated questions themselves can be noisy, i.e. nonsensical or ambiguous (see Table \ref{fluency_rating} under \S\ref{sec:res_ana}). 
To address this issue, we propose a simple training procedure that treats the synthetic and original data differently.
We first train a model on the synthetic data (\textbf{Synthetic Training}), then further train on the original, human-authored training set (\textbf{Organic Training}). 
The motivation is to correct any unfavorable noise that may have been learnt during the first stage, by subsequently training on original data as more recent training data is favored by neural models \cite{Goodfellow2014AnEI} . 

We also experiment with a mixing approach that minimizes a weighted average of the loss for the synthetic data and the original data, with an importance weight to downweight the synthetic examples to mitigate noise.
We find that two-stage training performs better than the importance-weighted loss
(see Section \ref{sec:analysis}).

\section{Experiments}
\label{ref:sec-exp}

\begin{table*}[]
\small
\centering
\begin{tabular}{lllllll}
              & \begin{tabular}[c]{@{}l@{}}CSQA\\ (Acc)\end{tabular} & \begin{tabular}[c]{@{}l@{}}\winogrande\\ (AUC)\end{tabular} & \begin{tabular}[c]{@{}l@{}}\codah\\ (Acc)\end{tabular} & \begin{tabular}[c]{@{}l@{}}HellaSwag-2K\\ (Acc)\end{tabular} & 
\begin{tabular}[c]{@{}l@{}}\textbf{Average} \end{tabular} \\ \toprule
\roberta (reported)            & 72.1                                                 & 66.4                                                       & -                                                     & -                                                                           &-                                       \\ 
\roberta (ours)               & 71.6                                                 & 67.5                                                       & 82.3                                                  & 75.4                                                              &74.2                                              \\ 
\backt     & 70.2                                                 & 67.2                                                       & 81.8                                                  & 73.0                                                                    &73.1                                        \\ 
\midrule[0.03em]
\gdaug-Rand      & 71.8                                                 & 70.9                                                       & 83.6                                                  & 75.9                                                             &75.6                                              \\ 
\gdaug-Influence & 72.1                                                 & 70.9                                                       & \textbf{84.3}                                                  & 75.8                                                                 &75.8                                         \\ 
\gdaug-Diversity & 72.3                                                 & 71.2                                                       & 83.5                                                  & 76.1                                                                &75.8                                            \\ 
\gdaug-Combo     & \textbf{72.6}                                                 &\textbf{71.4}                                                       & 84.0                                                  & \textbf{76.8}                                                           &\textbf{76.2}                                               \\ 
\bottomrule
\end{tabular}
\caption{Results on the test sets of four commonsense benchmarks. 
\roberta (reported) is the result for the \roberta-large baseline reported on public leaderboards.\footnotemark \roberta (ours) is re-evaluation of the \roberta-large model using our setup. 
All \gdaug\ methods outperform the baseline methods, and \gdaug-Combo performs the best overall. 
}
\label{test_res}
\end{table*}
 \footnotetext{\url{https://leaderboard.allenai.org/winogrande/submissions/public}, \url{https://www.tau-nlp.org/csqa-leaderboard}}
We present experiments on four commonsense multiple choice QA benchmarks: \comqa \cite{talmor-etal-2019-commonsenseqa}, \winogrande \cite{sakaguchi2019winogrande}, \codah \cite{Chen2019CODAHAA} and HellaSwag \cite{zellers2019hellaswag}.  Our techniques are also directly applicable to other closed-book multiple choice QA setups, such as science QA, and to textual entailment tasks with minor modifications.  To evaluate \gdaug 's extensibility to these settings, we also experiment with a textual entailment task, SNLI \cite{bowman2015large},
and a closed-book version of the ARC-Challenge Scientific QA task \cite{Clark2018ThinkYH} in which access to the scientific corpus for the ARC dataset (or any other information sources) is disallowed during test.
We simulate low-resource settings on the large HellaSwag and SNLI datasets by downsampling these to 2K and 3K training samples respectively; the other data sets are either already low-resource or have a low-resource component. 
Dataset details are provided in Appendix \ref{app:datasets}.

\paragraph{Robustness Evaluation}
In addition to measuring in-distribution performance, we also analyze robustness to perturbed or adversarial data.
Following \citet{wei-zou-2019-eda}, we perform WordNet-based \cite{WordNet} synonym replacement on the validation or test set (when test labels are available) with a $10\%$ replacement rate.\footnote{\url{https://github.com/jasonwei20/eda\_nlp}}
Our second evaluation with TextFooler \cite{jin2019bert}
identifies the most important words and replaces these with the most semantically and grammatically correct substitutes, until the model prediction is altered.
We adopt two metrics to measure robustness under TextFooler's attacks: 1) \textbf{failure rate}: the proportion of examples for which TextFooler fails to change the prediction and 2) \textbf{average perturbation ratio}: the average fraction of words replaced when TextFooler succeeds in altering a prediction. 
We re-implement TextFooler with two minor changes: we only swap words in questions, not answers, and we replace the Universal Sentence Encoder with  S\roberta\cite{Reimers2019SentenceBERTSE}.

\begin{table*}[]
\small
\centering
\begin{tabular}{lllllll}
             & CSQA & \winogrande & \codah & HellaSwag-2K  & \textbf{Average} \\ 
\toprule
\roberta (ours)               & 69.9 & 63.8       & 74.7  & 63.2                 &67.9                 \\ 
\backt     & 69.0   & 62.3       & 75.5  & \textbf{65.4}                 &68.1                  \\ 
\midrule[0.03em]
\gdaug-Rand      & \textbf{72.1} & 65.5       & 75.9  & 64.1                 &69.4                 \\ 
\gdaug-Influence & 71.0   & 65.7       & \textbf{76.2}  & 64.3                 & 69.3                 \\ 
\gdaug-Diversity & 71.6 & \textbf{66.0}       & 76.0  & 64.8                 & 69.6                 \\ 
\gdaug-Combo     & 72.0   & \textbf{66.0}       & 76.0  & 65.2                 & \textbf{69.8}               \\ 
\bottomrule
\end{tabular}
\caption{Results on WordNet-based synonym replacement sets. For \codah and HellaSwag-2K, we perturb test sets, as the labels are available. 
\gdaug-Combo achieves the highest average score. }
\label{syn_res}
\end{table*}

\subsection{Experimental Settings}
\label{ref:sec-exp:settings}
We use \roberta \cite{liu2019roberta} as our pretrained task model, and \gpt \cite{radford2019language} as our pretrained generator.\footnote{We used the HuggingFace library \cite{Wolf2019HuggingFacesTS}.}
We use validation performance to decide whether to do relabeling for \comqa and \winogrande, and apply relabeling by default on all other tasks (tuning this choice may boost performance).
To perform a controlled comparison, we restrict the synthetic set size to be equal across all methods. 
We repeat all experiments with 10 random restarts and pick the best model based on validation performance.
Additional experimental details, with hyperparameters, are provided in Appendix  \ref{app:hyp}.

\paragraph{Baselines}
Our first baseline is a finetuned \roberta model with no augmentation. 
We compare with existing work on data augmentation via a \backt approach from \citet{xie2019unsupervised}; under our setting the original and backtranslated data are mixed at random.\footnote{\url{https://github.com/google-research/uda/}}

\subsection{In-Distribution Results}
\label{sec:res_ana}

Our main results for commonsense question answering are reported in Table \ref{test_res}. 
All \gdaug variants outperform the baselines, highlighting the impact of generative data augmentation.
On average, every other variant achieves higher test performance than \gdaug-Rand, which further highlights the importance of our data selection approaches. In addition, influence and diversity selection methods score similarly, however, their combination (in \gdaug-combo) outperforms either alone, which suggests that they are complementary selection approaches.
More specifically, \gdaug-Combo performs the best on 3/4 tasks and obtains the highest average score. 
Further, \gdaug-Combo provides a 5.0\% absolute gain over previously published state-of-the-art results on \winogrande.\footnote{These results are state-of-the-art for our model class; higher scores have been obtained using a T5 model with roughly an order of magnitude more parameters than ours.} 
For \comqa, \gdaug-Combo outperforms the previous non-ensemble state-of-the-art \cite{Zhu2019FreeLBEA} by 0.4\%. 
We also achieve a new state-of-the-art on \codah, where the previous best (\bert-based) score was 67.5\% \cite{Chen2019CODAHAA}.
We find that \backt hurts performance, and uniformly underperforms the \roberta baseline.  See Appendix \ref{app:val} for validation set results.

\subsection{Robustness Results}

Table \ref{syn_res} presents our evaluation on synonym replacement sets. 
The \gdaug\ variants outperform the baselines, and \gdaug-Combo obtains the best average performance. 
Table \ref{text_fooler} shows results on the TextFooler adversarial attacks. Models trained with data augmentation are more robust to adversarial attacks, as all \gdaug\ variants and \backt outperform the \roberta baseline on both metrics.
\gdaug-Diversity obtains the best failure rate and average perturbation ratio (higher is better, in both metrics), and 
\gdaug-Combo performs comparably with slightly worse numbers. %
Overall, the findings suggest that optimizing diversity increases robustness.

\begin{table*}[]
\small
\centering
\begin{tabular}{lllllll}
              & CSQA & \winogrande & \codah & Hellaswag-2K & \textbf{Average} \\ 
\toprule
\roberta (ours)               & 14.8 / 12.6 & 4.5 / 7.8       & 30.9 / 15.8  & 17.4 / 9.8               & 16.9 / 11.5                 \\ 
\backt     & \textbf{17.0} / 12.9   & 5.0 / 8.2       & \textbf{37.1} / 15.9  & 20.2 / 10.2              & 19.8 / 11.8                \\ 
\midrule
\gdaug-Rand      &15.6 / \textbf{13.0}  &5.7 / 8.4        &36.2 / 15.9  & 20.0 / 10.6         & 19.4 / \textbf{12.0}                \\ 
\gdaug-Influence &16.3 / 12.8    & 5.4 / 8.4       & 34.9 / 15.8  & 19.2 / \textbf{10.7}            &19.0 / 11.9                  \\ 
\gdaug-Diversity &16.0 / 12.9 & \textbf{5.9} / 8.4      & 36.1 / \textbf{16.2}  & \textbf{21.4} / 10.4             & \textbf{19.9} / \textbf{12.0}                 \\ 
\gdaug-Combo     & 16.5 / 12.6   & \textbf{5.9} / \textbf{8.5}      & 35.2 / 15.7  & 21.3 / 10.5             &19.7 / 11.8                \\ 
\bottomrule
\end{tabular}
\caption{Robustness to TextFooler-based adversarial attacks (failure rate / average perturbation ratio, higher is better for both). 
Models trained with augmented data are more robust to TextFooler's attacks compared to models without data augmentation. On average, \gdaug-Diversity performs the best. }
\label{text_fooler}
\end{table*}

\subsection{Results on ARC and SNLI}

We explore \gdaug's applicability outside of the commonsense domain in Table \ref{arc_snli_table}, via evaluation on the closed-book ARC-Challenge Scientific QA. 
Valid science questions are hard to generate because their semantics need to be precise, and we find that many of \gdaug 's generations for ARC are noisy.
Perhaps surprisingly, nonetheless \gdaug\ outperforms the baselines by a large margin. 
\gdaug-Influence achieves the best in-distribution performance, while \gdaug-Diversity is the most robust against TextFooler but has worse accuracy than \gdaug-Rand. 
This may suggest that optimizing for quality is more important when the synthetic data is noisier.
\begin{table*}[]
\small
\centering
\begin{tabular}{llllll|llllll}
                                & \multicolumn{5}{c}{ARC-Challenge Scientific QA} &  \multicolumn{6}{c}{SNLI-3K}                        \\ \bottomrule
                                & Val.   & Test   & Syn.   & TF:Fail   & TF:Pert  & Val. & Test & Syn. & TF:Fail & TF:Pert & NLI Diag. \\ \toprule
RoBERTa (ours)                         & 43.5   & 39.4   & 35.2   & 6.6       & 9.3      & 91.8 & 88.6 & 77.5 & 17.0    & 20.2    & 56.7      \\
Backtranslation                 & 43.1   & 43.1   & 42.4   & 6.6       & 9.3      & 91.2 & 8.1  & \textbf{81.0} & 18.8    & \textbf{21.7}    & 54.0      \\ \toprule
\gdaug-Rand      & 50.8   & 48.1   & 43.4   & 12.9      & 10.8     & 91.8 & \textbf{89.0} & 78.6 & 17.7    & 20.6    & 57.4      \\
\gdaug-Influence & \textbf{51.5}   & \textbf{48.5}   & \textbf{45.2}   & 12.4      & \textbf{11.0}    & \textbf{92.3} & 88.7 & 78.6 & 18.0    & 20.7    & 56.9      \\
\gdaug-Diversity & 49.5   & 47.5   & 42.2   & \textbf{13.9}      & 10.8     & 92.0 & \textbf{89.0} & 79.4 & \textbf{19.0}    & 20.5    & \textbf{57.7}      \\
\gdaug-Combo     & 50.8   & 48.2   & 43.8   & 13.1      & 10.7     & 91.9 & 88.7 & 78.7 & 16.7    & 20.5    & 57.6     
\end{tabular}
\caption{Results on closed-book ARC-Challenge Scientific QA and SNLI-3K, along with robustness to synonym replacement, TextFooler (TF) attacks and NLI Diagnostics.  \gdaug improves accuracy and robustness. }
\label{arc_snli_table}
\end{table*}

We also evaluate \gdaug\ on a textual entailment using the SNLI dataset \cite{bowman2015large} in Table \ref{arc_snli_table}. 
This task has a different format; it is a pair-wise classification task with 3 labels (details in Appendix \ref{app:datasets}). 
We find that \gdaug\ slightly improves accuracy and robustness over baselines. The performance is likely affected by a label skew introduced by influence-based filtering.

\section{Analysis and Discussion}
\label{sec:analysis}
We now analyze \gdaug's performance, focusing on \winogrande where \gdaug\ offers the most benefit. We first identify several factors that affect performance, and then present evidence that \gdaug\ works by transferring knowledge from the pretrained generator to the task model.

\subsection{Factors that Affect \gdaug's Performance}

\paragraph{\gdaug\ is effective at different training sizes.}
\begin{figure}[t!]
\centering
  \includegraphics[width=0.92\columnwidth]{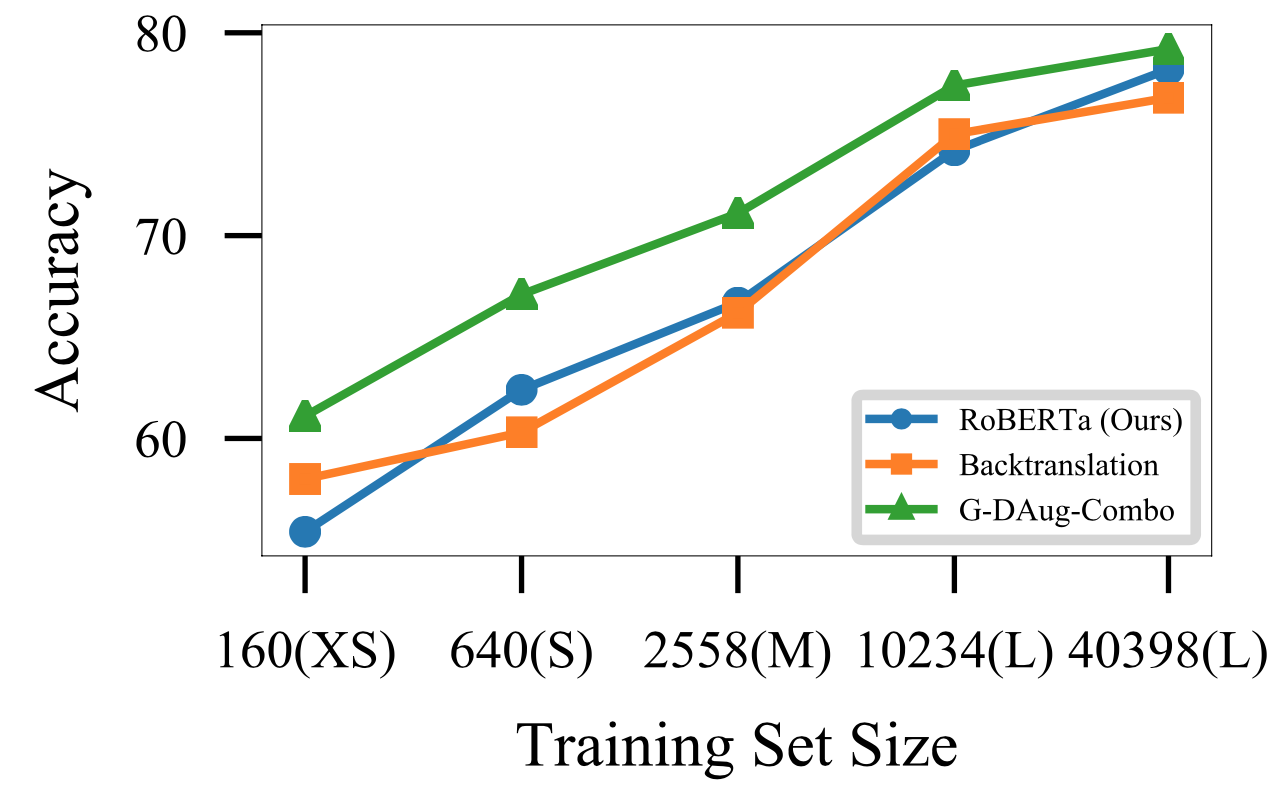}
  \caption{Validation results for different training set sizes on the \winogrande dataset (in log scale).  \gdaug helps more for smaller training sizes.}
  \label{fig:winogrande-lc}
\end{figure}
Figure \ref{fig:winogrande-lc} illustrates that our winning strategy, \gdaug-Combo, remains effective as the amount of training data varies, for \winogrande.
The improvement over baseline is largest in the low-resource (small training size) regime.
For the smallest sizes, XS and S, \gdaug-Combo increases the \textit{effective training size} by a factor of 4 (i.e. training on XS or S matches unaugmented \roberta's performance on S or M, respectively). 
In contrast, \backt only helps for the XS size, but hurts performance on larger sizes.

\paragraph{Staged training is essential.} 
\gdaug\ uses a two-staged training method (Section \ref{sec:data-aug:training}) aimed at mitigating the effect of noise in the generated data. 
We analyze alternative training protocols on the \winogrande-L dataset: {\em Mixing} (training on the union of generated and original data) and Importance Weighted Loss. Compared to a no-augmentation baseline (with accuracy of 75.9), two stage training (+1.8 increase) outperforms both mixing (+0.0) and importance weighted loss (+0.7).

\paragraph{Filtering synthetic data does not hurt accuracy.}

\begin{table}[hbt!]
\centering
\small
\begin{tabular}{lllll}
                     & Random &Influence &Diversity &Whole Pool \\ \toprule
 Size    &127478   &127478 &127478  &380700  \\ 
Acc &71.7 &74.4 &73.0 &73.1    \\ 
\bottomrule
\end{tabular}
\caption{Results comparing \gdaug's filtering methods against using the entire synthetic data pool for augmentation, on \winogrande-M. 
} %
\label{pool}
\end{table}
\gdaug 's filtering methods are designed to identify a high-quality and diverse subset of the generated data, to reduce training cost (compared to training on the entire generated pool) without harming accuracy.  We evaluate whether \gdaug\ is successful at achieving this in Table \ref{pool}, by comparing \gdaug\ against using the entire synthetic data pool for \gdaug-Influence and \gdaug-Diversity.\footnote{\gdaug-Combo utilizes a larger pool, so it is not comparable.} The selection approaches provide comparable or better accuracy compared to using the entire pool, despite using three times less data.

\subsection{Why Does \gdaug\ Work?}

Below, we present analysis suggesting that \gdaug\ works by transferring knowledge from the pretrained model to the task model.  In particular, we find that using a pre-trained generator is critical, and that the generated questions are often coherent, include new semantic units, and carry informative labels.

\paragraph{Using a {\em Pretrained} Generator is critical.} We analyze the impact of the pretrained generator by comparing our standard \gdaug-Rand setting with a setting where the generator is {\em not} pretrained, but instead trained from scratch. We find that using \gpt trained from scratch results in a score of $67.8\%$ on the \winogrande-M validation set. This is a slight improvement (by $0.2\%$) over the unaugmented baseline, but is far inferior to the $3.9\%$ improvement obtained when using the pretrained \gpt.
This suggests that using a pretrained generator is critical for \gdaug.

\paragraph{Synthetic data labels are important.}

\begin{table}[hbt!]
\small
\centering
\begin{tabular}{lll}
   & \winogrande-L      &CSQA \\ \toprule
Baseline        &75.9           &77.1 \\ 
\midrule[0.03em]
Generator label  & 76.2     &\textbf{78.1}                       \\ 
Random relabeling    & 66.8  &77.1                          \\ 
Model relabeling & \textbf{77.7}        & 77.7    \\ \bottomrule
\end{tabular}
\caption{Validation accuracy of \gdaug with different labeling methods on \winogrande-L and \comqa. 
Random labels hurt accuracy, and model relabeling helps on \winogrande but not on \comqa .}
\label{relabel}
\end{table}
Even fully unsupervised language model pretraining can improve performance, when using task-relevant data \cite{domains2020}.  
This raises the question of whether \gdaug\ boosts performance by simply exposing the model to more task-relevant text, or if the generated labels are in fact informative. 
A related question is whether \gdaug 's optional self-supervised relabeling improves performance. 
We analyze these questions for \winogrande-L and \comqa in Table \ref{relabel}, evaluating \gdaug\ with three labeling methods: (i) generator labels, (ii) random relabeling, and (iii) relabeling with a task model. 
When the generator labels are flipped randomly, \gdaug\ is unable to outperform the baselines for either dataset (in fact, it dramatically underperforms on \winogrande-L). 
This implies that the correctness of labels is crucial for \gdaug.
Self-supervised relabeling provides a 1.5\% absolute gain in \winogrande-L, but a 0.4\% drop in \comqa, which suggests its utility is task-dependent.

\begin{table*}[]
\small
\center
\begin{tabular}{c|l|l|c|c}
\textbf{Rating} & \textbf{Description}           & \textbf{Examples} & \textbf{Count} & \textbf{Pct.}\\ \hline \hline
1               & Nonsensical                     & 
    \begin{tabular}[c]{@{}l@{}}
        What is a square leg made of made out of?\\
        What country does a cow go to make a milk run?
    \end{tabular}
    & 54  & 3.89\% \\ \hline
2               & Ambiguous or unanswerable       & 
    \begin{tabular}[c]{@{}l@{}}
        A person is a human, but they are called what?\\ 
        He hated flying, the controls were what?
    \end{tabular}
    & 306 & 22.06\% \\ \hline
3               & Minor errors (e.g., grammar) &
    \begin{tabular}[c]{@{}l@{}}
        What do you put on your head to do when you're swimming?\\
        Where does a bugle call be played?
    \end{tabular}
    & 138 & 9.95\%        \\ \hline
4               & Coherent and Fluent            & 
    \begin{tabular}[c]{@{}l@{}}
        What is a person likely to feel when applying for jobs?\\ 
        If you're running late for work what would you be doing?        %
    \end{tabular}
    & 889 & 64.10\%    \\
\end{tabular}
\caption{Examples and prevalence of generated commonsense questions with different manually-assigned fluency ratings, for the \comqa dataset. Ratings of 3 and higher correspond to questions that are answerable and address common sense, and most of \gdaug 's generated questions fall into this category.}
\label{fluency_rating}
\end{table*}

\begin{figure}[htb!]
\centering
  \includegraphics[width=0.9\columnwidth]{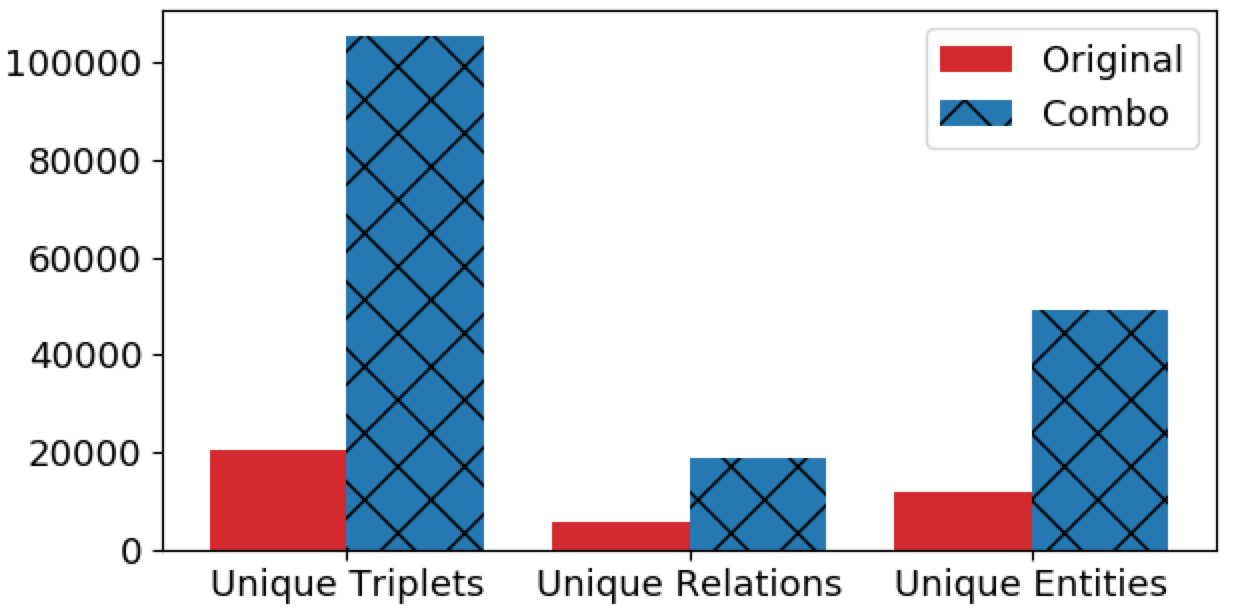}
  \caption{OpenIE analysis on the original data and synthetic data used by \gdaug-Combo on \winogrande-M. The synthetic dataset contains many more unique semantic units compared to the original dataset. %
  }
  \label{fig:openie}
\end{figure}

\paragraph{\gdaug\ introduces new semantic units.}
We investigate {\em how distinct} the generated questions are from each other and from the original training data.
We observe that \gdaug\ only rarely
generates exact duplicate questions  (e.g., on \comqa, 0.06\% of the questions are duplicates).
We further investigate if \gdaug\ introduces new entities and relations to the training data, or if it merely reuses the ones found in the original training set.  
We quantify the diversity of our synthetic dataset compared to the original data by counting the number of unique semantic units produced by performing Open Information Extraction \cite{banko2007} on the data. %
Specifically, we run the Stanford Open IE package \cite{Angeli2015LeveragingLS} and report the number of unique triplets, relations and entities extracted from our \winogrande-M datasets in Figure \ref{fig:openie}.  
The synthetic data includes many more unique semantic units than the original training data, suggesting that \gdaug\ does introduce new semantic units in the training set.  %

\paragraph{\gdaug produces mostly fluent questions.} 
To evaluate \gdaug 's output for fluency, we employ three human annotators to rate generated \comqa questions for their coherence and answerability on a scale of 1 to 4, where a rating of 3 denotes an acceptable question.
We obtained a total of 1,387 labels.  
We measured annotator agreement on a separate set of 50 questions, obtaining a Fleiss' Kappa of 0.41, which is at the low end of moderate annotator agreement, acceptable given the subjective nature of the task. 
A large ($74.04\%$) majority of questions met the acceptability threshold, with an overall average rating of 3.34.
Examples are shown in Table \ref{fluency_rating}. 

Next, we ask annotators to answer the 1,027 acceptable questions, where they can edit choices (but not questions) if they are unable to pick a unique correct answer from the given choices. 
The editing rate is relatively high, at 55.3\%. We mix these human-labeled examples with the original training set to train a \roberta model, and obtain $78.1\%$ validation accuracy, which is comparable to \gdaug, despite using approximately 50x fewer questions. 
This suggests that human labels can provide higher leverage than the noisy labels from \gdaug, although human labeling is expensive.

Additional analyses, provided in Appendix \ref{app:additional}, show that model sharpness approximated by the Hessian trace \cite{Yao2019PyHessianNN} %
does not completely explain \gdaug's performance; and, \gdaug\ is more effective than ensembling with a finetuned generator. 
\section{Related Work}
\label{sec:related_work}
Data augmentation is a common practice in computer vision, where it takes the form of image transformations like translation and rotation  \cite{perez2017effectiveness}. 
For language tasks, data augmentation is less straightforward. 
Broadly, previous augmentation methods have used back-translation architectures \cite{sennrich-etal-2016-improving, xie2019unsupervised}, heuristics based on syntactic and semantic properties of text including word replacements using a thesaurus \cite{zhang2015character,wei-zou-2019-eda} and word embeddings \cite{wang-yang-2015-thats, fadaee-bisazza-monz:2017:Short2, kobayashi-2018-contextual, wu2019conditional}, and recently, generative models for synthesizing novel examples for text classification and reading comprehension \cite{AnabyTavor2019NotED, Kumar2020DataAU, puri2020training}. 
Our framework is similar to the last of these as we focus on generative models for data augmentation, but our work is the first to present a generative approach for the challenging commonsense QA setting, and we introduce new data selection approaches to improve the informativeness and diversity of synthetic data. 

Concurrently, there has been work on generating adversarial examples for analyzing black-box classifiers. These approaches use generative adversarial networks \cite{zhao2018generating} and population-based optimization algorithms \cite{alzantot-etal-2018-generating}. Previous work has also presented methods to generate questions for reading comprehension \cite{heilman-smith-2010-good, Rus2012, alberti-etal-2019-synthetic,Puri2020TrainingQA}, online tutoring \cite{lindberg-etal-2013-generating}, factual QA \cite{serban-etal-2016-generating} and visual question generation \cite{mostafazadeh-etal-2016-generating}. A comprehensive survey on neural question generation can be found in \citet{pan2019recent}. Our work is distinct in that it targets question generation in a closed-book setting, investigates the generation of \textit{answers} as well as \textit{distractors}, and is aimed at data augmentation.

\section{Conclusion}
\label{sec:conclusion}

We introduced \gdaug, a novel data augmentation framework to generate synthetic training data, preserving quality and diversity. %
We demonstrate that \gdaug\ is effective on multiple commonsense reasoning benchmarks, with improvements on in-distribution performance, as well as robustness against perturbed evaluation sets and challenge sets. 
Our analysis shows that \gdaug\ tends to perform better in low-resource settings and that our data selection strategies are important for performance.
Future work might explore more sophisticated methods to enhance quality and diversity of generated training data, including having humans-in-the-loop for relabeling.

\section*{Acknowledgments}
This work was supported in part by NSF Grant IIS-1351029.  We thank Iz Beltagy, Jonathan Bragg, Isabel Cachola, Arman Cohan, Mike D'Arcy, Daniel King, Kyle Lo, and Lucy Lu Wang for helpful comments. %

\bibliography{anthology,emnlp2020}
\bibliographystyle{acl_natbib}
\clearpage
\newpage
\clearpage
\appendix

\section{Datasets}
\label{app:datasets}

\paragraph{CommonsenseQA} \cite{talmor-etal-2019-commonsenseqa}: CommonsenseQA is a multiple choice QA dataset that consists of 12,247 examples, which aims to test commonsense reasoning capabilities. We use the official random split 1.11 which is an 80/10/10
split. We apply greedy decoding to generate answers, as answers are fairly short for this dataset. 
\paragraph{\winogrande} \cite{sakaguchi2019winogrande}: \winogrande is a benchmark for commonsense reasoning, inspired by the original Winograd Schema Challenge design \cite{Levesque2011TheWS}, with a larger dataset size and higher difficulty level. It consists of 44K questions with five different training sizes: 160, 640, 2,558, 10,234 and 40,398 questions. The evaluation metric is Area Under the (learning) Curve. We observe that applying top-2 greedy decoding on the answer generator is able to yield a satisfactory set of choices, so the distractor generator is not used in this task.  The Winograd schema requires that questions in twin pairs have opposite labels \cite{Levesque2011TheWS}.  We use the following method to generate twin questions: 1. generate a sequence until a blank symbol "\_" is produced. 2. use two independent runs of sampling to complete the question in two different ways to form twins. The above process does not guarantee that the labels will differ for the two twins, so we further filter out generated pairs that do not have different labels.
\paragraph{\codah} \cite{Chen2019CODAHAA}: \codah is an adversarially-constructed benchmark which tests commonsense reasoning using sentence-completion questions, inspired by the Swag dataset \cite{zellers2018swagaf}. It contains 2,801 questions in total, and uses 5-fold cross validation for evaluation.\footnote{The original CODAH work does not specify a particular 5-fold split, so we choose these randomly.  We will release our splits for replicability.} We lower the temperature to 0.5 for the answer generation in order to increase the confidence of the generated answers. 
\paragraph{HellaSwag} \cite{zellers2019hellaswag}: HellaSwag is a more challenging version of the Swag dataset \cite{zellers2018swagaf}, and the task is similar to \codah. The dataset consists of 70K questions where each question comes from one of two domains: ActivityNet or WikiHow. In order to test our methods under a low-resource setting, we downsample the training set to 2,000 examples. We take a random sample of 1000 questions from the original validation set to serve as our validation data, and another non-overlapping random sample of 5,000 questions from the same set as our test data. The generation settings are the same as \codah's.

\paragraph{SNLI} \cite{bowman2015large}: SNLI is a natural language inference dataset with 570K pairs of labeled sentences. The label assigned to each sentence pair is one of entailment, contradiction or neutral. For low-resource experiments, we downsample the dataset to 3K training examples, which contains 1K unique premises and a hypothesis for all three labels. Similarly, we use a downsampled development set with 999 examples (333 premises and 3 hypotheses for each label). The generative model is fine-tuned by providing the premise, label and hypothesis, separated by special delimiters marking the beginning and end of each element.

\paragraph{ARC-Challenge} \cite{Clark2018ThinkYH}: The ARC Dataset consists of 7,787 natural grade-school science questions that are used on standardized tests. The ARC-Challenge Set contains 2,590 questions answered incorrectly by both a retrieval-based algorithm and a word co-occurence algorithm. We use the official split, which has 1,119 train, 299 validation, and 1,172 test examples. The generation settings are the same as \comqa's.

\section{Validation Set Results}
In Table \ref{val_res}, we summarize our main results on the validation sets, comparing the \gdaug methods against an unaugmented baseline and a backtranslation augmentation baseline. All \gdaug methods consistently outperform the baseline methods in every benchmark. The proposed selection methods provide an extra boost on average, compared to \gdaug-Rand. Among those, \gdaug-Influence achieves the best performance across all tasks, which is expected as \gdaug-Influence selects examples which are helpful in reducing validation loss. Interestingly, \gdaug-Combo scores lower than \gdaug-Influence, although it outperforms \gdaug-Diversity. Finally, backtranslation does not demonstrate any benefit and obtains lower results compared to the augmented baseline in all benchmarks. 

\label{app:val}
\begin{table*}[]
\small
\centering
\begin{tabular}{lllllll}
Method              & \begin{tabular}[c]{@{}l@{}}CSQA\\ (Acc)\end{tabular} & \begin{tabular}[c]{@{}l@{}}\winogrande\\ (AUC)\end{tabular} & \begin{tabular}[c]{@{}l@{}}\codah\\ (Acc)\end{tabular} & \begin{tabular}[c]{@{}l@{}}HellaSwag-2K\\  (Acc)\end{tabular} & \begin{tabular}[c]{@{}l@{}}\textbf{Average} \end{tabular} \\ \toprule
\roberta (reported)            & 78.4                                                 & 66.6                                                      & -                                                     & -                                                            &-          &-                                               \\ 
\roberta (ours)              & 77.1                                                 & 68.4                                                       & 84.2                                                  & 75.2                                                              &76.2                                              \\ 
Backtranslation     & 76.4                                                 & 67.7                                                       & 83.4                                                  & 74.2                                                               &75.4                                                \\ \midrule
\gdaug-Rand      & 78.1                                                 & 72.0                                                       & 85.7                                                  & 77.2                                                                 &78.3                                        \\ 
\gdaug-Influence & \textbf{78.8}                                                 & \textbf{73.0}                                                       & \textbf{87.2}                                                  & \textbf{78.3}                                                           &\textbf{79.3}                                                  \\ 
\gdaug-Diversity & 78.1                                                 & 72.8                                                       & 86.0                                                  & 76.6                                                                      &78.4                                     \\ 
\gdaug-Combo     & 78.2                                                 & 72.7                                                       & 86.7                                                  & 77.5                                                               &78.8\\ \bottomrule
\end{tabular}
\caption{Results on the validation sets of four commonsense benchmarks. 
All \gdaug methods outperform the baseline methods, in particular, \gdaug-Influence performs the best on all tasks, which is expected as it selects examples which are helpful in reducing validation loss.
}
\label{val_res}
\end{table*}

\section{Hyperparameter Settings and Input Formats }
\label{app:hyp}
Hyperparameter settings for finetuning GPT-2, \roberta and \gdaug are shown in Tables \ref{app:gpt2_finetune_hyperparams},  \ref{app:gpt2_finetune_hyperparams_wino}, \ref{app:original_hyperparams}, \ref{app:original_hyperparams_wino} and \ref{app:gdaug_hyperparams}. We manually tune the learning rate and the number of epochs for GPT-2 finetuning based on validation perplexity. For finetuning \roberta baseline models, we select the number of epochs from \{1,3,5,8,10\} based on validation accuracy for CSQA, \winogrande and HellaSwag-2K. For \codah, SNLI-3K and ARC-Challenge, we simply use 5 epochs. For \gdaug synthetic training, we train all models using a learning rate of 5e-6 for one epoch. For \gdaug organic training, we use the same hyperparameter settings as \roberta baselines (except for CSQA and HellaSwag-2K, where we find reducing 2 epochs gives significantly better results). In Tables \ref{app:gen_input_format} and \ref{app:task_input_format}, we specify the input formats for finetuning GPT-2 and \roberta.  Finally, we benchmark the running time of our implementations of the influence and diversity selection methods on the task of selecting 127,478 examples from a pool consisting of 380,700 candidates for \winogrande-M. We use one Nvidia 2080 Ti GPU and one Intel Core I9-7900X with 10 cores and a clockspeed of 3.3 GHz. The running time of the influence and diversity algorithms is about 8.3 hours and 2.9 hours, respectively.

\begin{table*}[]
\centering
\footnotesize
\begin{tabular}{ll}
\bf Task                     & \bf Format \\ \toprule
CSQA                 & $ \text{Q: Where can I stand on a river to see water falling without getting wet? A: waterfall }	\langle /s 	\rangle  $          \\
\winogrande &$\langle /s \rangle \text{Feeling a draft, William asked Neil to please close the front door because \_ was closer.} \langle /s \rangle \text{Neil} \langle /s 	\rangle $   \\ 
\codah                     & $\langle /s \rangle \text{I am always very hungry before I go to bed. I am} \langle /s \rangle \text{concerned that this is an illness.}  \langle /s \rangle $    \\ 
HellaSwag-2K               & $\langle /s \rangle \text{A man is on a sandy beach, playing croquette. he} \langle /s \rangle \text{is parasailing, making a random move.}  \langle /s \rangle$     \\ 
SNLI-3K  & $\langle \text{PREM} \rangle \text{Five black dogs run in a field.} \langle \text{/PREM} \rangle \langle \text{ANS} \rangle \text{entailment}  \langle \text{/ANS} \rangle \langle \text{HYP} \rangle \text{Some animals running.}  \langle \text{/HYP} \rangle$\\
ARC-Challenge &$ \text{Q: Which of the following is an example of a physical change? A: breaking a glass }	\langle /s 	\rangle  $ \\
\bottomrule
\end{tabular}
\caption{ Input formats for GPT-2. "Q:" and "A:" are the prefix for a question and a candidate answer (choice). }
\label{app:gen_input_format}
\end{table*}

\begin{table*}[]
\centering
\footnotesize
\begin{tabular}{ll}
\bf Task                     & \bf Format \\ \toprule
CSQA                 & $ \langle s 	\rangle  \text{Q: Where can I stand on a river to see water falling without getting wet?}	\langle /s 	\rangle \text{ A: waterfall }	\langle /s 	\rangle  $          \\
\winogrande &$\langle s \rangle \text{Feeling a draft, William asked Neil to please close the front door because \_ was closer.} \langle /s \rangle \text{Neil} \langle /s 	\rangle $   \\ 
\codah                     & $\langle s \rangle \text{I am always very hungry before I go to bed. I am} \langle /s \rangle \text{concerned that this is an illness.}  \langle /s \rangle $    \\ 
HellaSwag-2K               & $\langle s \rangle \text{A man is on a sandy beach, playing croquette. he} \langle /s \rangle \text{is parasailing, making a random move.}  \langle /s \rangle$     \\ 
SNLI-3K &  $\langle s \rangle \text{Five black dogs run in a field.} \langle /s \rangle \text{Some animals running.}  \langle /s \rangle$ \\
ARC-Challenge &$ \langle s 	\rangle  \text{Q: Which of the following is an example of a physical change?}\langle /s 	\rangle \text{A: breaking a glass }	\langle /s 	\rangle  $ \\
\bottomrule
\end{tabular}
\caption{ Input formats for \roberta. "Q:" and "A:" are the prefix for a question and a candidate answer (choice). }
\label{app:task_input_format}
\end{table*}

\begin{table*}[t]
\begin{center}
\footnotesize
\begin{tabular}{lcccccc}
\toprule
\bf Hyperparam  & \bf CSQA &\bf \winogrande & \bf \codah &\bf HellaSwag-2K &\bf SNLI-3K &\bf ARC-Challenge \\
\midrule 
Version  & Large  &Medium  &Medium  &Medium  &Large &Medium\\
Hardware &I9-7900X &RTX 2080Ti &RTX 2080Ti  &RTX 2080Ti &RTX 8000 &RTX 2080Ti \\
Optimizer &AdamW  &AdamW  &AdamW &AdamW &AdamW &AdamW\\\
Adam $\beta_1$ & 0.9  &0.9 &0.9  &0.9 &0.9 &0.9\\
Adam $\beta_2$  & 0.98 &0.98 &0.98  &0.98 &0.999 &0.98\\
Adam $\epsilon$ &1e-6 &1e-6   &1e-6   &1e-6 &1e-8 &1e-6 \\
Mixed Precision  &No  &Yes  &Yes    &Yes  &Yes &Yes\\
LR (q/a/d) &1e-5/5e-6/2e-5 &* &4e-5/5e-5/5e-5 &4e-5/5e-5/5e-5  &5e-5 &2e-5/1e-5/1e-5\\
Epochs (q/a/d) & 3/5/3 &*  & 3/3/3 &3/3/3 &3 &3/5/5\\
Grad Clipping &1.0 &1.0   &1.0 &1.0 &1.0 &1.0\\
Weight Decay & 0.01 & 0.01 & 0.01 &0.01 &0.0 &0.01 \\
Batch Size & 16 & 16  & 16 &16 &16 &16 \\
Max Length (q/a/d) &62/70/70 &72/72/-   &62/92/92  &62/128/128 &128 &90/120\\
Warmup Ratio & 0.06 & 0.06 & 0.06  &0.06 &0.06 &0.06\\
LR Decay & Linear &Linear &Linear &Linear &Linear &Linear  \\
\bottomrule
\end{tabular}
\end{center}
\caption{
Hyperparameter settings for finetuning GPT-2. "q/a/d" stands for "question/answer/distractor". Some hyperparameters for \winogrande is shown in a separate table as they vary with the train size.   
}
\label{app:gpt2_finetune_hyperparams}
\end{table*}

\begin{table*}[t]
\begin{center}
\footnotesize
\begin{tabular}{lccccc}
\toprule
\bf Hyperparam  & \bf XS  & \bf S &\bf M &\bf L &\bf XL \\
\midrule 

LR (q/a) &5e-5/5e-5 &2e-5/5e-5  &2e-5/5e-5 &2e-5/5e-5 &1e-5/5e-5  \\
Epochs (q/a) &8/12  & 6/6 &3/3 &3/3 &3/1\\

\bottomrule
\end{tabular}
\end{center}
\caption{
Hyperparameter settings for finetuning GPT-2 on \winogrande.
}
\label{app:gpt2_finetune_hyperparams_wino}
\end{table*}

\begin{table}[htbp!]
\centering
\footnotesize
\begin{tabular}{ll}
                    & Test AUC \\ \toprule
Baseline      & 67.5     \\ 
Baseline + Generator                      & 67.5 \\ 
\gdaug-Combo                      & \textbf{71.4}\\ 
\bottomrule
\end{tabular}
\caption{Test performance of an unaugmented baseline model and the same model ensembled with a finetuned GPT-2 generator on \winogrande. We use weighted average ensemble with weights tuned on validation data.  
}
\label{ensemble}
\end{table}

\begin{table*}[t]
\begin{center}
\footnotesize
\begin{tabular}{lcccccc}
\toprule
\bf Hyperparam  & \bf CSQA & \bf \winogrande & \bf \codah &\bf HellaSwag-2K &\bf SNLI-3K &\bf ARC-Challenge\ \\
\midrule 
Version &Large &Large &Large &Large &Large &Large\\ 
Hardware &RTX 2080Ti &RTX 2080Ti &RTX 2080Ti  &RTX 2080Ti &RTX 8000 &RTX 2080Ti \\
Optimizer &AdamW    &AdamW  &AdamW &AdamW &AdamW &AdamW\\\
Adam $\beta_1$ & 0.9   &0.9 &0.9 &0.9 &0.9 &0.9\\
Adam $\beta_2$  & 0.98 &0.98  &0.98 &0.98 &0.98 &0.98\\
Adam $\epsilon$ &1e-6  &1e-6  &1e-6 &1e-6 &1e-6 &1e-6 \\
Mixed Precision  &Yes  &Yes   &Yes    &Yes  &Yes &Yes\\

LR &1e-5 &* &1e-5 &1e-5  &1e-5 &1e-5\\
Epochs & 5 &* & 5 &3 &5 &5\\
Grad Clipping &0.0   &0.0 &0.0 &0.0 &0.0 &0.0\\
Weight Decay & 0.01 & 0.01 & 0.01 &0.01 &0.01 &0.01 \\
Batch Size & 16 & 16  & 16 &16 &16 &16 \\
Max Length &70  &70  &90  &128 &128 &120\\
Warmup Ratio & 0.06 & 0.06 & 0.06 &0.06 &0.06 &0.06\\
LR Decay & Linear &Linear & Linear &Linear &Linear &Linear  \\
\bottomrule
\end{tabular}
\end{center}
\caption{
  Hyperparameter settings for finetuning \roberta. Some hyperparameters for \winogrande are shown in a separate table as they vary with the training set size. 
}
\label{app:original_hyperparams}
\end{table*}

\begin{table*}[t]
\begin{center}
\footnotesize
\begin{tabular}{lccccc}
\toprule
\bf Hyperparam  & \bf XS  & \bf S &\bf M &\bf L &\bf XL \\
\midrule 

LR &1e-5  &1e-5 &1e-5  &1e-5 &1e-5\\
Epochs & 10 & 8 &5 &5 &5\\

\bottomrule
\end{tabular}
\end{center}
\caption{
Hyperparameter settings for finetuning \roberta on \winogrande.
}
\label{app:original_hyperparams_wino}
\end{table*}

\begin{table*}[t]
\begin{center}
\footnotesize
\begin{tabular}{lcccccc}
\toprule
\bf Hyperparam  & \bf CSQA & \bf \winogrande  & \bf \codah &\bf HellaSwag-2K &\bf SNLI-3K &\bf ARC-Challenge \\
\midrule 
Synthetic Data Size &50K &$\sim$ 50K-130K\footnotemark &100K  &50K &100K &50K \\
LR (synthetic) &5e-6  &5e-6 &5e-6  &5e-6 &5e-6 &5e-6\\
Epochs (synthetic) & 1 & 1 &1 &1 &1 &1\\

\bottomrule
\end{tabular}
\end{center}
\caption{
Additional hyperparameter settings for \gdaug Two-Stage Training. For finetuning on the original data, we use the same settings as \roberta (except for CSQA and HellaSwag-2K, where we find reducing 2 epochs gives significantly better results).
For Winogrande, we generate 400K examples before the rejection procedure (see Appendix \ref{app:datasets}). The examples retained after the rejection procedure approximately ranges from 50K-130K depending on the training size.}
\label{app:gdaug_hyperparams}
\end{table*}

\section{Influence Functions}
\label{sec:influence_extra}

In practice, since the generalization error is usually approximated by validation loss, a training example $x_i$ is considered detrimental if it increases validation loss, i.e.:
\begin{align}
&\mathcal{L}(\mathcal{X},\boldsymbol{\theta})=\frac{1}{|\mathcal{X}|}\sum_{x\in\mathcal{X}}l(x,\boldsymbol{\theta}),\\
\label{val_d}
&\mathcal{L}(\mathcal{X}_{val},\hat{\boldsymbol{\theta}}(\mathcal{X}_{train}\cup \{x_i\}))-\mathcal{L}(\mathcal{X}_{val},\hat{\boldsymbol{\theta}}(\mathcal{X}_{train}))>0,
\end{align}
where $\mathcal{X}_{train} = \{x_i\}_{i=1}^{N}$ is a training set,  $\mathcal{X}_{val} = \{x_i\}_{i=1}^{M}$ is a validation set, $l$ is a loss function, and $\hat{\boldsymbol{\theta}}(\mathcal{X}_{train})=  \underset{\boldsymbol{\theta}\in \Theta}{\mathrm{argmin}}\  \mathcal{L}(\mathcal{X}_{train},\boldsymbol{\theta})$ is an empirical risk minimizer.

The main result from previous work \citep{Atkinson1983ResidualsAI,Koh2017UnderstandingBP} tells us that the influence of upweighting a training example $x$ by some small $\epsilon$ on the model parameters $\hat{\boldsymbol{\theta}}$ with the corresponding parameter space $\Theta$ is given by:
\begin{align}
&\hat{\boldsymbol{\theta}}_{\epsilon,x}=\underset{\boldsymbol{\theta}\in \Theta}{\mathrm{argmin}}\ \epsilon l(x,\boldsymbol{\theta}) + \frac{1}{\sum_{i=1}^{N} w_i}  \sum_{i=1}^{N} w_i l(x_i,\boldsymbol{\theta}) \\
\label{up_inf}
&\mathcal{I}_{up,params}(x):= \evaluat*{\frac{d\hat{\boldsymbol{\theta}}_{\epsilon,x}}{d\epsilon}}_{\epsilon=0}=-H^{-1}_{\hat{\boldsymbol{\theta}}}\nabla_{\boldsymbol{\theta}}l(x,\hat{\boldsymbol{\theta}}),
\end{align}
where $w_i$ is weight for the training example $x_i$ and $H_{\hat{\boldsymbol{\theta}}}= \frac{1}{\sum_{i=1}^{N} w_i} \sum_{i=1}^{N} w_i\nabla_{\boldsymbol{\theta}}^{2}l(x_i,\hat{\boldsymbol{\theta}})$ is the Hessian evaluated at $\hat{\boldsymbol{\theta}}$. The above result is a slight generalization of \citet{Koh2017UnderstandingBP}, since the simple average used in that work is a special case of our weighted average, but it is straightforward to generalize their proof to our weighted empirical risk case and we omit the details of the proof in this paper. Then, we apply the chain rule to get the influence of upweighting $x$ on the validation loss:
\begin{align}
\label{up_inf_loss}
&\mathcal{I}_{up,loss}(x):= \evaluat*{\frac{d\mathcal{L}(\mathcal{X}_{val},\hat{\boldsymbol{\theta}}_{\epsilon,x})}{d\epsilon}}_{\epsilon=0}\\
&= \nabla_{\boldsymbol{\theta}}\mathcal{L}(\mathcal{X}_{val},\hat{\boldsymbol{\theta}})^\top \mathcal{I}_{up,params}(x).
\end{align}
Note that $\mathcal{L}(\mathcal{X}_{train},\boldsymbol{\theta})$ can be rewritten as the following weighted average form to incorporate a new training example $x_{new}$:\\
\begin{align*}
\label{eq:avg_inf}
    \mathcal{L}(\mathcal{X}_{train},\boldsymbol{\theta}) =& \frac{1}{\sum_{i=1}^{N+1}w_i}\sum_{i=1}^{N+1} w_i l(x_i,\boldsymbol{\theta}),
 \end{align*}
 where $ w_i = 1 \forall{i \neq N+1}$, $w_{N+1} = 0$ and $x_{N+1} = x_{new}$. 
Adding the new training example $x_{new}$ is equivalent to upweighting $x_{N+1}$ by $\frac{1}{N}$:

\begin{align*}
&\mathcal{L}(\mathcal{X}_{train}\cup \{x_{new}\},\boldsymbol{\theta})=\frac{N}{N+1}(\frac{1}{N}l(x_{N+1},\boldsymbol{\theta})\\
&+\frac{1}{\sum_{i=1}^{N+1}w_i}\sum_{i=1}^{N+1}w_{i}l(x_i,\boldsymbol{\theta}))\\
&\propto\frac{1}{N}l(x_{N+1},\boldsymbol{\theta}) + \frac{1}{\sum_{i=1}^{N+1}w_i}\sum_{i=1}^{N+1}w_{i}l(x_i,\boldsymbol{\theta}). 
\end{align*}

Applying the influence function $\mathcal{I}_{up,loss}(x)$, we obtain the following linear approximation of the validation loss change upon adding the training example $x_{new}$:
\begin{align}
&\mathcal{L}(\mathcal{X}_{val},\hat{\boldsymbol{\theta}}(\mathcal{X}_{train}\cup \{x_{new}\}))-\mathcal{L}(\mathcal{X}_{val},\hat{\boldsymbol{\theta}}(\mathcal{X}_{train}))\\
&\approx \frac{1}{N}\mathcal{I}_{up,loss}(x_{new}). 
\end{align}
We adopt the stochastic estimation method described in \citet{Koh2017UnderstandingBP} to efficiently compute $\mathcal{I}_{up,loss}$. Detrimental synthetic data will have $\frac{1}{N}\mathcal{I}_{up,loss}>0$.

\section{Diversity Selection using Embedding Distance}
\label{app:emb}
We define our embedding distance based diversity measure as the sum of the cosine distances between every pair of selected examples. To attempt to maximize this measure, we use a greedy algorithm that at each iteration randomly samples 10K candidate examples from the pool, and selects the candidate that maximizes the distance between it and its nearest neighbor in the set of examples selected so far. We use S\roberta \cite{Reimers2019SentenceBERTSE} as our sentence embedding method and Faiss \cite{JDH17} as our nearest neighbor searcher. We compare the embedding distance based measure with the unigram approach on \winogrande dataset. The embedding distance based diversity selection is not found to be more effective than the unigram approach, in fact it performs 0.6\% worse. 

\section{Additional Analysis}
\label{app:additional}

\paragraph{Sharpness Analysis.} 
Previous work \cite{Hochreiter1997FlatM,Keskar2016OnLT,Yao2019PyHessianNN} has shown that models with flatter local minima tend to generalize better. Moreover, \citet{Hao2019VisualizingAU} show that pretraining helps BERT to achieve flat and wide optima in the finetuning stage, which partially explains its performance benefits. We investigate whether \gdaug 's data augmentation may also encourage flatter optima.  Specifically, using the fact that a larger Hessian trace for a model implies a sharper local minimum \cite{Yao2019PyHessianNN}, we compute the Hessian trace of 10 baseline and 10 \gdaug-Combo methods using the Hutchinson Method \cite{Avron2011RandomizedAF} and find an average relative decrease of $9.5\%$ for \gdaug-Combo, suggesting that \gdaug does find slightly flatter optima. Likewise, when comparing the best performing models of each approach, \gdaug-Combo's best model is slightly flatter than the baseline (a relative decrease of $0.2\%$). However, we also find the contradictory fact that, over the 20 models, flatter optima tend to be associated with {\em worse} task performance (Spearman correlation of $0.39$, $p \approx 0.09$).  So, it does not appear that sharpness explains \gdaug's performance advantage over the baseline. 
A more thorough analysis of this hypothesis is an item of future work.

\paragraph{Generator/Task Model Ensemble.} 
\gdaug harnesses pretrained knowledge from \gpt in order to improve a \roberta -based task model.  A more standard approach for model combination (albeit, with twice the computational cost at runtime) would be to ensemble the two models instead.  We evaluate ensembling a baseline \roberta model with a finetuned \gpt generator for \winogrande in Table \ref{ensemble}. We adopt a weighted-average ensemble method, where the weights are tuned on validation data (the tuning is important to achieve peak performance). The ensemble model performs same as the baseline model, and \gdaug-Combo outperforms both of them by 3.9\%. This suggests that \gdaug is more effective than simply ensembling the finetuned generator.

\end{document}